\documentclass[9pt, conference]{IEEEtran}
\IEEEoverridecommandlockouts
\usepackage{cite}
\usepackage{amsmath,amssymb,amsfonts}
\usepackage{algorithmic}
\usepackage{graphicx}
\usepackage{textcomp}
\usepackage{xcolor}
\def\BibTeX{{\rm B\kern-.05em{\sc i\kern-.025em b}\kern-.08em
    T\kern-.1667em\lower.7ex\hbox{E}\kern-.125emX}}

\usepackage{tikz}
\usepackage{calc}

\usepackage[noabbrev]{cleveref}

\usepackage{booktabs}
\usepackage{mathtools}
\usepackage{amsmath}
\usepackage{multirow} 
\usepackage{makecell}
\usepackage{times}
\usepackage{epsfig}
\usepackage{fancyhdr}
\usepackage[normalem]{ulem}
\usepackage{soul}
\usepackage{xcolor}
\usepackage{dblfloatfix}

\newcommand{\CJ}[1]{{\color{black}{#1}}}

\usepackage{pifont}
\newcommand\mynuma[1]{\ifcase#1 \or \ding{172}\or \ding{173}\or
  \ding{174}\or \ding{175}\or \ding{176}\or \ding{177}%
  \or \ding{178}\or \ding{179}\or \ding{180}\or \ding{181}\else *\fi\relax}
\newcommand\mynumb[1]{\ifcase#1 \or \ding{182}\or \ding{183}\or
  \ding{184}\or \ding{185}\or \ding{186}\or \ding{187}%
  \or \ding{188}\or \ding{189}\or \ding{190}\or \ding{191}\else *\fi\relax}

\usepackage{enumitem}
\usepackage{booktabs}
\usepackage[flushleft]{threeparttable}
\usepackage{etoolbox}
\usepackage{lipsum}
\usepackage{wrapfig}

\usepackage{caption}

\appto\TPTnoteSettings{\footnotesize}

\newcommand{\name}{Instant-NeRF}

\definecolor{Note_color}{rgb}{1.0, 0.0, 0.0}

\begin{document}

\title{
{Instant-NeRF: \underline{Instant} On-Device \underline{Ne}ural \underline{R}adiance \underline{F}ield Training via Algorithm-Accelerator Co-Designed Near-Memory Processing}
}

\author{
\IEEEauthorblockN{Yang (Katie) Zhao$^1$, Shang Wu$^2$, Jingqun Zhang$^1$, Sixu Li$^1$, Chaojian Li$^1$, Yingyan (Celine) Lin$^1$}
\IEEEauthorblockA{$^1$\textit{Georgia Institute of Technology}, $^2$\textit{Rice University}}
\IEEEauthorblockA{\textit{\{eiclab.gatech, jzhang3368, sli941, cli851, celine.lin\}@gatech.edu, \{sw99\}@rice.edu}}
}

\maketitle

\begin{abstract}
Instant on-device Neural Radiance Fields (NeRFs) are in growing demand for unleashing the promise of immersive AR/VR experiences, but are still limited by their prohibitive training time. Our profiling analysis reveals a memory-bound inefficiency in NeRF training. To tackle this inefficiency, near-memory processing (NMP) promises to be an effective solution, but also faces challenges due to the unique workloads of NeRFs, including the  random hash table lookup, random point processing sequence, and heterogeneous bottleneck steps. Therefore, we propose the first NMP framework, {\name}, dedicated to enabling instant on-device NeRF training. Experiments on eight datasets consistently validate the effectiveness of {\name}.


\end{abstract}

\section{Introduction} 
\label{sec:intro}

3D scene reconstruction
is crucial for numerous Augmented and Virtual Reality (AR/VR) applications~\cite{workrooms}. 
Neural radiance fields (NeRFs)~\cite{mildenhall2020nerf,muller2022instant} have yielded state-of-the-art (SOTA) rendering quality.
Therefore, many researchers have tried to speed up NeRF training toward instant NeRF-based 3D reconstruction in many emerging AR/VR applications. Despite the success achieved in accelerating NeRF training on cloud GPUs~\cite{muller2022instant}, NeRF-based 3D reconstruction on edge devices~\cite{quest_pro} is still not feasible.
To close the aforementioned gap between the desired instant on-device 3D scene reconstruction and the currently achievable NeRF training efficiency on edge devices, we first conduct extensive profiling measurements of the SOTA efficient NeRF training method, iNGP~\cite{muller2022instant}, on a SOTA edge GPU, XNX~\cite{xnx}, to identify the bottlenecks. 
Specifically, iNGP represents a 3D scene with a multi-resolution hash table of trainable embedding vectors, followed by two small multi-layer perceptions (MLPs) for capturing the density and RGB colors, respectively.
Our profiling analysis reveals that 
computing the embedding vectors and 
executing the MLPs mentioned above are the efficiency bottlenecks.
Furthermore, we identify 
that these bottlenecks are caused by the bounded bandwidth of dynamic random-access memory (DRAM). Specifically, 
the memory bandwidth utilization is 5.24$\times$$\sim$21.44$\times$ higher than the corresponding Floating-Point Unit/Arithmetic-Logic Unit (FPU/ALU) utilization. The causes of this memory-bound inefficiency are that (1) the random hash table lookup requires a high \textbf{memory bandwidth} to fetch embedding vectors and (2) both the hash table and intermediate data of the MLPs require a much larger \textbf{memory capacity} than that of the on-chip cache capacity. 

To overcome the aforementioned 
bottlenecks, emerging near-memory processing (NMP) architectures~\cite{devic2022pim, kwon2019tensordimm, asgari2021fafnir} are promising solutions.
This is because they can provide higher memory bandwidth by integrating computation logic units closer to the memory. For example, recent works 
deploy computation logic units at the bank level in DRAM and achieve around 10$\times$ peak bandwidth improvement~\cite{devic2022pim}. Additionally, their per-bank memory capacity can be as large as hundreds of megabytes (MB); it thus can provide sufficient on-chip memory for NeRF training.

\begin{figure}[t]
  \centering
  \includegraphics[width=1.0\linewidth]{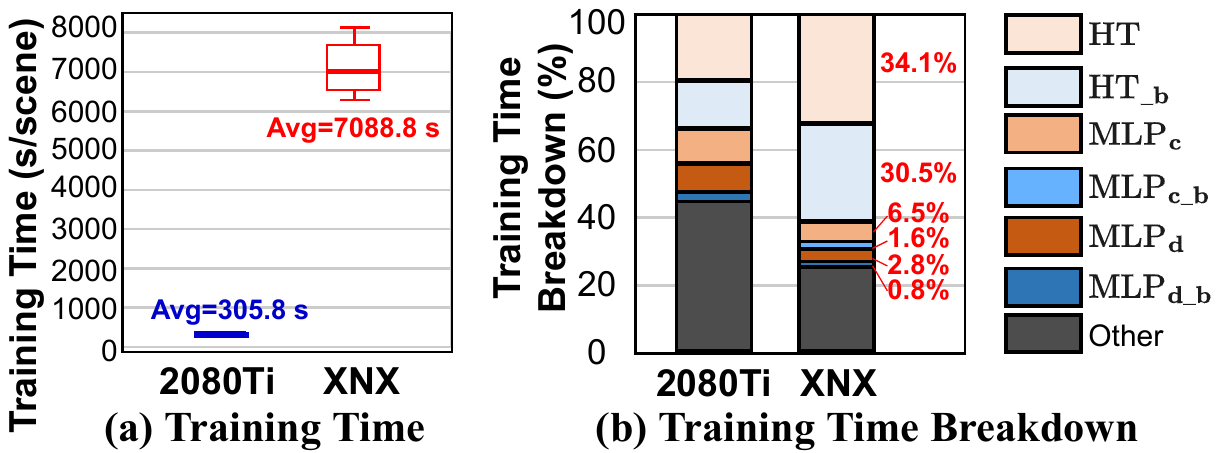}
\caption{(a) Training time and (b) its breakdown, when running the SOTA efficient NeRF training method~\cite{muller2022instant} on a cloud GPU (2080Ti~\cite{2080ti}) and an edge GPU (XNX~\cite{xnx}). Here HT and $\text{HT}_{\_\mathbf{b}}$ denote the hash table accesses and the corresponding back-propagation, $\text{MLP}_{\mathbf{c}}$ and $\text{MLP}_{\mathbf{c}\_\mathbf{b}}$ denote the MLP processing for the color features and the corresponding back-propagation, and $\text{MLP}_{\mathbf{d}}$ and $\text{MLP}_{\mathbf{d}\_\mathbf{b}}$ denote the MLP processing for the density features and the corresponding back-propagation, respectively. (See more details in Sec.~\ref{sec:iNGP_background}.)}
  \vspace{-1.0em}
  \label{fig:iNGP_comp}
\end{figure}



Despite their promise in alleviating the bottlenecks of NeRF training, directly applying NMP architectures to train iNGP~\cite{muller2022instant} would not be efficient due to the following \textbf{three challenges}. \textbf{First}, the required random hash table lookups in iNGP can result in reduced effective memory bandwidth for NMP architectures. This is because the memory requests adopt a row-wise granularity (e.g., 1KB (kilobytes)~\cite{devic2022pim,oh20143lpddr4}), whereas each hash table entry (i.e., one embedding vector) only uses 32 bits. Furthermore, the random hash table lookups can cause bank conflicts if two memory requests access the same bank with different addresses, leading to serialized computations and increased latency.
\textbf{Second}, the random processing sequence of points in a 3D scene can lead to non-sequential accesses to the same hash table entries, and thus incur long-latency memory accesses. 
\textbf{Third}, there exist heterogeneous bottleneck steps (e.g., index calculation via hash mapping function, hash table lookup, and MLP) as well as varying data types (e.g., integer 32-bit (INT32), floating-point 32-bit (FP32)) in iNGP, which require dedicated support. 

To address the identified bottlenecks hindering instant on-device NeRF training, we make the following contributions:

\label{sec:preliminary_iNGP}
\begin{figure}[!t]
  \centering
  \includegraphics[width=0.96\linewidth]{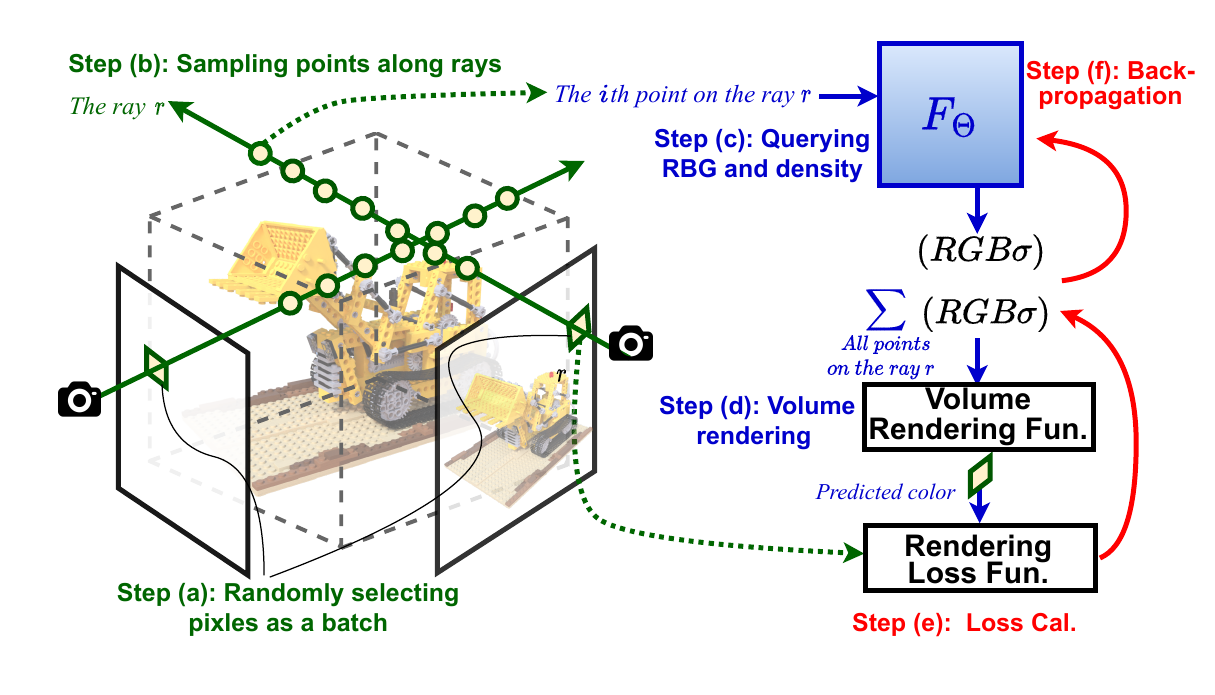}
  \vspace{-1.2em}
    \caption{An illustration of vanilla NeRFs'~\cite{mildenhall2020nerf} training process.}
  \label{fig:NeRF_training_process}
    \vspace{-1.em}
\end{figure}

\begin{itemize}[leftmargin=*]
    \item We conduct extensive profiling measurements of the SOTA efficient NeRF training method~\cite{muller2022instant} on SOTA edge devices over eight datasets, and identify the corresponding memory-bound efficient bottlenecks (Sec.~\ref{sec:background}). Our profiling results can inspire future innovative NeRF training techniques.
    
    \item We propose {\name}, an algorithm-accelerator co-design framework, to tackle the challenges of leveraging the promising NMP architecture to alleviate the memory-bound bottlenecks in NeRF training process. To the best of our knowledge, {\name} is the first to leverage an NMP architecture for achieving instant on-device NeRF-based 3D reconstruction. 
    \item Our {\name} algorithm (Sec.~\ref{sec:alg}) integrates a locality-sensitive 3D hash mapping function to map neighboring vertices in a 3D scene to neighboring hash table entries to tackle the memory bandwidth bottleneck and adopts a ray-first point streaming order to enhance the local register hit rates and reduce required memory access requests. 
    \item Our {\name} accelerator (Sec.~\ref{sec:arch}) integrates a dedicated mapping scheme optimized for {\name}'s algorithm and a mixed-precision computation logic to cope with different involved data types. Furthermore, we propose a heterogeneous inter-bank parallelism design, orchestrating the different computation and memory patterns in the heterogeneous bottleneck steps with the inter-bank parallelism opportunities while minimizing the inter-bank data movement overhead. 
    
    
    \item Comprehensive experiments (Sec.~\ref{sec:evaluation}) show that {\name} provides up to 266.1$\times$ speedup over SOTA edge GPU baselines while maintaining a similar rendering quality. 
\end{itemize}

\section{Background and Motivation}
\label{sec:background}
\subsection{iNGP with SOTA NeRF Training Efficiency}
\label{sec:iNGP_background}

\textbf{Vanilla \CJ{NeRFs'} Training Pipeline and Cost.} Given images from sparsely sampled views of a scene, NeRFs learn to reconstruct the scene to generate images \CJ{from} any arbitrary view \cite{mildenhall2020nerf}. Fig.~\ref{fig:NeRF_training_process} shows vanilla NeRFs' training process, involving six steps. Specifically,
\textbf{Step~(a)} randomly selects pixels from the input images as a batch, where selected pixels' coordinates and viewing directions serve as NeRFs' inputs with their RGB colors being the corresponding ground truth labels during training; In \textbf{Step~(b)}, for each \CJ{selected} pixel, multiple 3D points \CJ{are sampled along the ray that is formulated as $\mathbf{r}=\mathbf{o}+t\mathbf{d}$ ($t \in \{t_i\}, i \in [1, N]$). 
Here $\mathbf{o}$ is the coordinate of the camera's position, $\mathbf{d}$ is the unit vector that \CJ{points to the pixel from $\mathbf{o}$}, $N$ is the total number of the sampled points along each ray, and $\{t_i\}$ denotes the set of the distance between $\mathbf{o}$ and point $\mathbf{o}+t_i\mathbf{d}$;} 
In \textbf{Step~(c)}, given the $i$\CJ{-}th point on the ray $\mathbf{r}$, the corresponding spatial location \CJ{$\mathbf{o}+t_i\mathbf{d}$} and direction \CJ{$\mathbf{d}$} are applied to an MLP model ($F_\Theta$ in Fig.~\ref{fig:NeRF_training_process}), which then outputs the color $\mathbf{c}_i$ and density $\sigma_i$ of this sampled point;  
\textbf{Step~(d)} synthesizes each pixel's color via volume rendering~\cite{max1995optical}:
\vspace{-0.5em}
\begin{align}
\hat{\mathbf{C}}(\mathbf{r}) = \sum_{i=1}^{N} T_{i}(1-\exp(-\sigma_i (t_{i+1}-t_{i})))\mathbf{c}_i
\label{eq:volume_redering}
\end{align}

\noindent where $T_i = \exp(-\sum_{j=1}^{i} \sigma_{j} (t_{j+1}-t_{j}))$;
\textbf{Step~(e)} calculates the loss
$\mathcal{L}=\sum_{\mathbf{r} \in \mathcal{R}}\left\|\mathbf{\hat{C}}(\mathbf{r})-\mathbf{C}(\mathbf{r})\right\|_2^2$,
where $\mathcal{R}$ is the \CJ{ray set of the current training batch} and $\mathbf{C}(\mathbf{r})$ is the \CJ{corresponding} ground truth color;
\textbf{Step~(f)} \CJ{does} back-propagation. 

As the MLP model \CJ{requires $>$}1 million FLOPs \CJ{per input point}, \CJ{vanilla NeRFs~\cite{mildenhall2020nerf} training typically require $>$1 day per-scene even on a SOTA cloud GPU~\cite{2080ti}.}

\begin{figure}[t]
  \centering
  \includegraphics[width=1.0\linewidth]{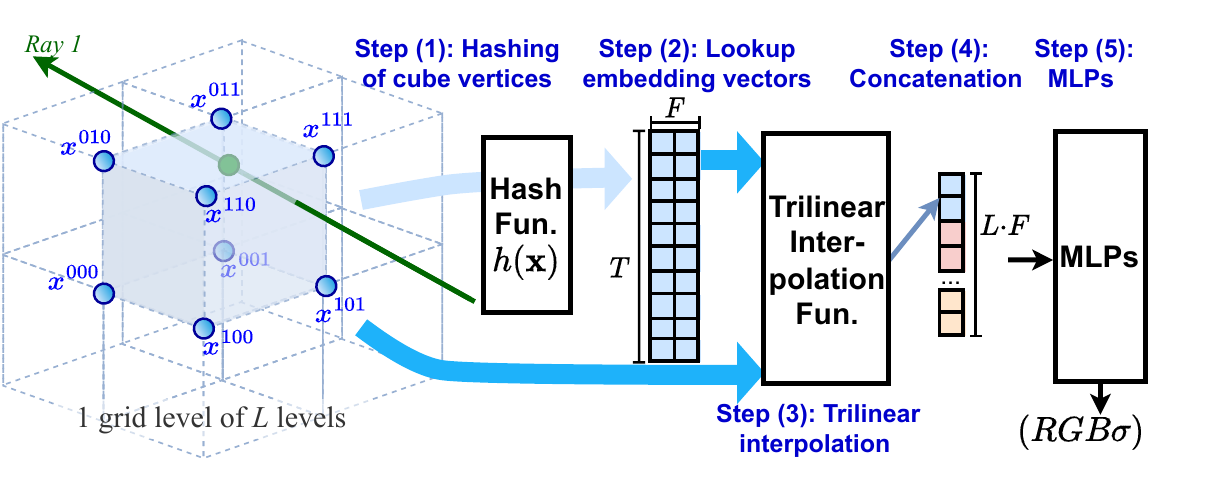}
  \vspace{-2.2em}
  \caption{How iNGP~\cite{muller2022instant} implements Step~(c) of vanilla NeRFs.}
  \label{fig:iNGP_training_process}
  \vspace{-0.5em}
\end{figure}

\textbf{iNGP's Training Pipeline.} To reduce the training cost of vanilla NeRFs, iNGP~\cite{muller2022instant} replaces the MLP model in 
the above dominant \textbf{Step~(c)} with a multi-resolution hash table of trainable embedding vectors and two much smaller MLPs, enabling efficient training (e.g., 305.8s/scene in Fig.~\ref{fig:iNGP_comp}(a)). Here the hash table encodes multi-resolution (i.e., a total of $L$ resolutions) grids \CJ{into 
$T$ vectors per level with each having a length of $F$}.
Fig.~\ref{fig:iNGP_training_process} shows how iNGP\cite{muller2022instant} implements \textbf{Step~(c)} of vanilla NeRFs with 
 five steps: \textbf{Step~(1)} - Hashing of cube vertices: Given an input point location $\mathbf{x}=(x_0,x_1,x_2)$, $L$ surrounding 3D
  cubes (one cube per level) are first found; 
\textbf{Step~(2)} - Lookup embedding vectors: Based on the surrounding cubes, iNGP fetches corresponding embedding vectors from the hash table, which has $T$ entries per level;
\textbf{Step~(3)} - Trilinear interpolation: Computing the embeddings of points at \CJ{each} level via trilinearly interpolating the embeddings of the corresponding eight surrounding vertices; 
\textbf{Step~(4)} - Concatenation: iNGP concatenates the resulting embeddings of all levels as the inputs of the subsequent MLP models;
\textbf{Step~(5)} - Execute MLPs: Generating the density and RGB color features via the above MLP models. \underline{In this work}, 
we use ``$\text{HT}$'' to denote \textbf{Steps~(1$\sim$3)} and ``$\text{MLP}_{\mathbf{d}}$''/``$\text{MLP}_{\mathbf{c}}$'' for the forward process of the density/color MLP in \textbf{Step~(5)}; 
The corresponding back-propagation that updates the embedding vectors and MLP parameters are denoted as ``$\text{HT}_{\_\mathbf{b}}$'' and ``$\text{MLP}_{\mathbf{c}\_\mathbf{b}}$''/``$\text{MLP}_{\mathbf{d}\_\mathbf{b}}$''. 

\subsection{Profiling SOTA Efficient NeRF Training Method on GPUs}
\label{sec:profling_analysis}

To understand the bottleneck of iNGP~\cite{muller2022instant} training, we first profile its training \CJ{process} on GPUs.  
\textbf{Profiling Setup.} \ul{Profiling Platform:} We use SOTA GPUs including two edge GPUs (XNX~\cite{xnx} and TX2~\cite{tx2}) and one cloud GPU (2080Ti~\cite{2080ti}). Tab.~\ref{tab:GPU_sepc} summarizes the device specifications of these GPUs as well as one SOTA edge GPU adopted by Meta's latest VR glass Quest Pro~\cite{quest_pro}. Since the \CJ{adopted} edge GPUs for profiling have a comparable on-chip cache \CJ{size} and \CJ{FP32/INT32/FP16 computation performance}, the profiling results can reflect the bottlenecks of NeRF-based on-device 3D reconstruction on VR/AR devices. \ul{Two-Stage Profiling Method:} Stage~(1) characterizes the runtime of each step (or kernel) in the training process to locate the dominant steps and Stage~(2) profiles the DRAM bandwidth utilization and computation resource utilization of the \CJ{located dominant steps} to identify the source of inefficiency. The GPU runtime and resource utilization are measured by NVIDIA's \textit{nvprof} toolbox. \ul{Algorithm \& Datasets:} We evaluate iNGP~\cite{muller2022instant} on eight datasets of Synthetic-NeRF~\cite{mildenhall2020nerf}. Each \CJ{dataset} takes 35,000 iterations with a batch size of 256K sampled points/iteration.

\begin{table}[t]
\centering
\caption{A summary of the considered SOTA GPUs' specs.} 
\begin{threeparttable}
\resizebox{\linewidth}{!}{
\begin{tabular}{c|c c c|c}
    \hline
    \multirow{2}{*}{\textbf{Spec.}} & \multirow{2}{*}{\textbf{XNX~\cite{xnx}}} & \textbf{Edge GPUs} & \multirow{2}{*}{\textbf{Quest Pro$^*$~\cite{quest_pro}}} & \textbf{Cloud GPU} \\
    &  & \textbf{TX2~\cite{tx2}} &  & \textbf{2080Ti~\cite{2080ti}} \\ \hline
    \textbf{Tech.} & 16nm & 16nm & 7nm & 12nm \\ \hline
    \textbf{Power} & 20W  & 15W  & 5W & 250W \\ \hline
    \multirow{3}{*}{\textbf{DRAM}}  & 128-bit 16GB & 128-bit 8GB & 64-bit 12GB & 352-bit 11GB \\ 
                           & LPDDR4$\times$   & LPDDR4   & LPDDR5   & GDDR6 \\ 
                           & 59.7GB/s & 25.6GB/s & 44.0GB/s  & 616GB/s \\ \hline
    \textbf{GPU L2 Cache} & 512KB & 512KB & 1MB & 5.5MB \\ \hline
    \textbf{FP32/INT32} & 885 GFLOPS  & 750 GFLOPS  & 955 GFLOPS  & 13.45 TFLOPS \\ \hline
    \textbf{FP16}       & 1.69 TFLOPS & 1.50 TFLOPS & 1.85 TFLOPS & 26.9 TFLOPS \\ \hline
    \textbf{Training Time} & 7088s/scene & 44653s/scene & N/A  & 306s/scene \\ \hline   
\end{tabular}
    }
\end{threeparttable}
\begin{tablenotes}
\footnotesize
\item{$^*$: Specs. of Qualcomm Adreno 650 GPU in Meta's Quest Pro VR glass~\cite{quest_pro}.}
\end{tablenotes}
\label{tab:GPU_sepc}
 \vspace{-1.em}
\end{table} 

\textbf{Profiling Result Analysis.} Although iNGP reduces \CJ{the training time on cloud GPUs to $<$6 minutes per scene, it still requires $>$1 hour per scene on the edge GPUs, as shown in Fig.~\ref{fig:iNGP_comp}(a).} From the training time breakdown in Fig.~\ref{fig:iNGP_comp}(b), we can locate \textbf{four} efficiency-bottleneck steps/kernels: \textbf{$\text{HT}$}, \textbf{$\text{HT}_{\_\mathbf{b}}$}, \textbf{$\text{MLP}_{\mathbf{d}}$}, and \textbf{$\text{MLP}_{\mathbf{c}}$}. Note that as the training on XNX is 2.9$\times$ faster than that on TX2, we only visualize \CJ{the profiling results on XNX} in Fig.~\ref{fig:iNGP_comp}. These steps/kernels (with their back-propagation \CJ{processes}) account for 76.4\% of the total training time. 
After locating the dominant steps/kernels, we measure their DRAM read/write throughput and FPU/ALU performance (i.e., FP or INT operations per second). Here the DRAM bandwidth utilization is calculated as \CJ{the portion of} the achieved DRAM throughput over the maximum bandwidth provided by the GPU. Similarly, we can calculate the computation resource utilization for FPU/ALU.

\begin{figure}[t]
  \centering
  \includegraphics[width=0.88\linewidth]{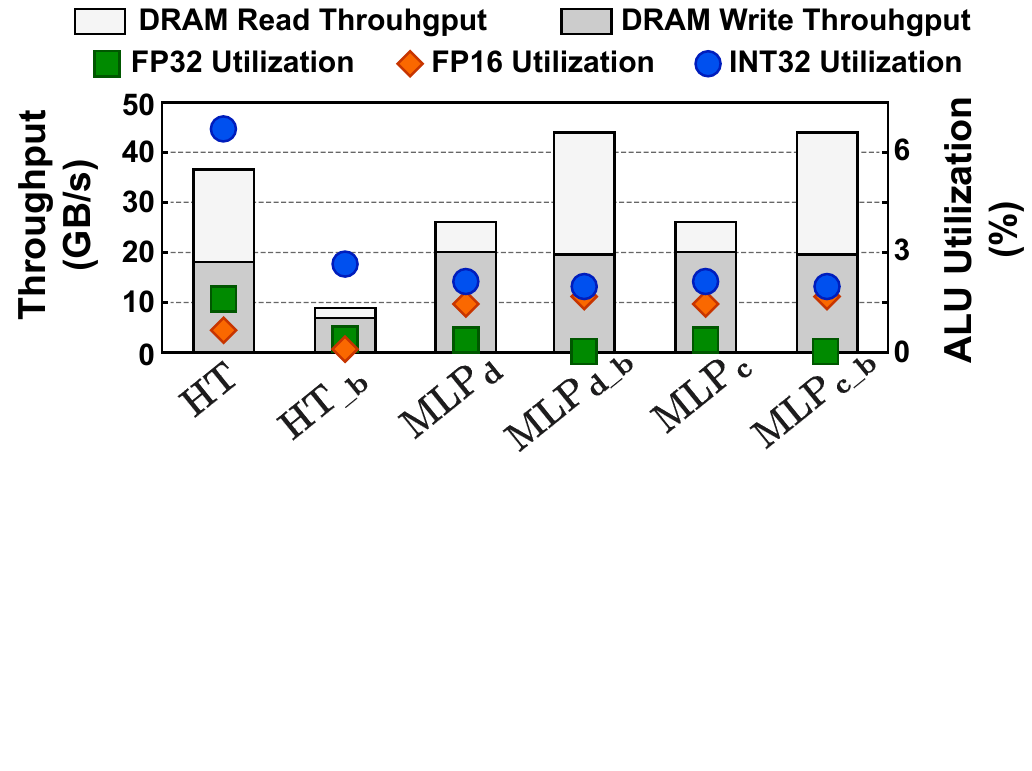}
  \vspace{-7.7em}
\caption{The DRAM read/write throughput and computation logic utilization of the \CJ{efficiency-bottleneck} steps/kernels (and their corresponding back-propagation \CJ{processes}) when running iNGP~\cite{muller2022instant} training method on \CJ{a} SOTA edge GPU~\cite{xnx}.}
    \label{fig:iNGP_kernel}
      \vspace{-1.em}
\end{figure}

Our profiling results show \CJ{the following} three observations: \ul{First}, the steps/kernels exhibit DRAM \textbf{bandwidth-bound bottleneck}, where the DRAM bandwidth utilization is 5.24$\times$$\sim$21.44$\times$ higher than the \CJ{FPU/ALU} utilization (see Fig.~\ref{fig:iNGP_kernel}). Specifically, $\text{HT}$/$\text{MLP}_{\mathbf{d}}$/$\text{MLP}_{\mathbf{d}\_\mathbf{b}}$/$\text{MLP}_{\mathbf{c}}$/$\text{MLP}_{\mathbf{c}\_\mathbf{b}}$ achieves 61.3\%/47.5\%/73.7\%/47.5\%/73.7\% DRAM bandwidth utilization (given the 59.7GB/s \CJ{maximum} DRAM bandwidth), \CJ{while the FP32/FP16/INT32 utilization of the five aforementioned steps/kernels is all $\leq$1.5\%/$\leq$1.6\%/$\leq$6.4\%, respectively.}
\CJ{Note that} both the DRAM and FPU/ALU utilization are relatively low for $\text{HT}_{\_\mathbf{b}}$, as $\text{HT}_{\_\mathbf{b}}$ involves frequent write-after-read operations to update the embedding vector gradients\CJ{, where idleness exists between the read and write operations.}
\ul{Second}, the causes of \CJ{the exhibited} memory-bound inefficiency \CJ{above} are (1) random lookups to the hash table, which stores multi-resolution grids' embedding vectors, requires a high \textbf{memory bandwidth} and (2) the on-chip GPU cache \textbf{memory capacity} is too small for handling the hash table storage \CJ{requirements} and processing the MLPs. Specifically, each individual level of the hash table is 2MB, which is 2$\times$$\sim$4$\times$ larger than the available edge GPU cache capacity, let alone the 64MB intermediate data for the MLP processing\CJ{, as suggested in Tab.~\ref{tab:data_size}}. \ul{Third}, the \CJ{the index calculation via hash mapping function~\cite{muller2022instant}}, an important part of the hash table lookups, consumes a large portion of the total INT32 ALU utilization. \CJ{Specifically, we observe that the INT32 ALU utilization, which is caused by the index calculation, is 4.2$\times$$\sim$160.7$\times$ higher than that of the FP32/FP16 utilization, which is caused by the computations of other steps/kernels.} This calls for dedicated architecture support for the index calculation in iNGP.

\begin{figure*}[t]
  \centering
  \begin{minipage}{.4\textwidth}
  \includegraphics[width=1.\linewidth]{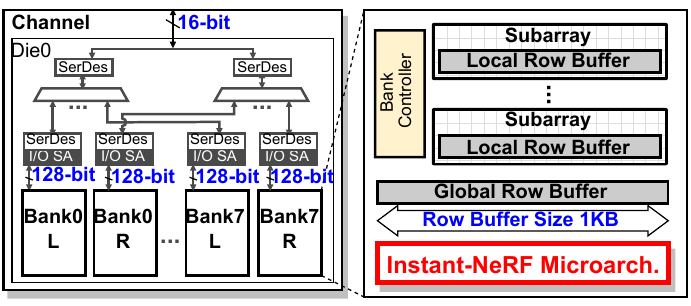}
    \caption{High-level LPDDR4~\cite{oh20143lpddr4} DRAM organization and {\name} microarchitecture's integration location at each bank.}
    \label{fig:lpddr}
\end{minipage} \hfill
  \begin{minipage}{.2\textwidth}
    \includegraphics[width=1.\linewidth]{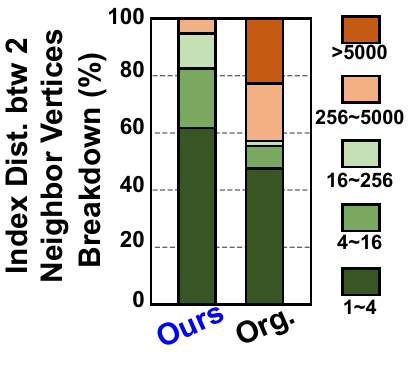}
    \caption{The breakdown of index distances between two neighboring vertices.}
    \label{fig:alg_hashing}
\end{minipage} \hfill
  \begin{minipage}{.35\textwidth}
\includegraphics[width=1.\linewidth]{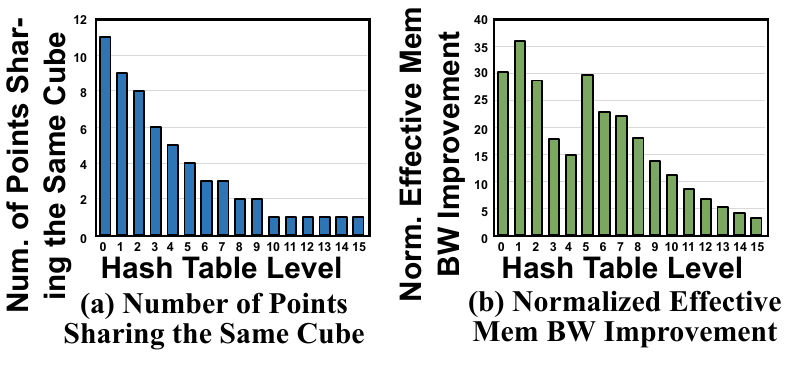}
\caption{(a) The number of points sharing the same cube and (b) the normalized effective memory bandwidth improvement thanks to the proposed algorithmic techniques.}
\label{fig:alg_stream}
\end{minipage}
\vspace{-1.em}
\end{figure*}
\subsection{Identified Opportunities for NMP-based NeRF Training}
\CJ{As analyzed in Sec.~\ref{sec:profling_analysis}, offloading the detected} memory-bound bottleneck steps to NMP architectures is promising in reducing \CJ{the total} training latency. 
We consider a type of DRAM widely used by edge devices~\cite{tx2,xnx}: Low Power Multiple Dual In Memory Module 4 (LPDDR4)~\cite{oh20143lpddr4}, as an example to discuss the opportunities of using NMP to accelerate \CJ{the training process}. As shown in Fig.~\ref{fig:lpddr}, 
an LPDDR4 channel typically has one rank with one die per rank; 
One LPDDR4 die contains 16 physical banks, which share a common I/O interface. While the I/O interface width is 16-bit and the internal prefetch structure has a width of 128-bit/physical bank (i.e., 16n prefetch structure~\cite{oh20143lpddr4}), the row buffer within each bank provides a data width of 1KB. This 
organization offers an intrinsic parallelism opportunity~\cite{devic2022pim} for addressing the \textbf{memory bandwidth} bottleneck in iNGP training. Second, for a typically adopted 8/16GB 128-bit LPDDR4 memory system in edge devices~\cite{tx2,xnx}, each bank has 128MB$\sim$256MB capacity, providing a sufficient \textbf{memory capacity} for \CJ{NeRFs} training. Finally, as each bank contains subarrays, different subarrays 
can be accessed mostly independently~\cite{kim2012case}.
Therefore, \CJ{our proposed} {\name} adopts a near-bank NMP architecture with subarray parallelism for enabling on-device NeRF training\CJ{, as illustrated in Sec.~\ref{sec:arch}}.



\section{{\name} Algorithm}
\label{sec:alg}
We introduce two algorithmic techniques in {\name}
to address the memory-bound bottlenecks of iNGP arising from (1) the need for random hash table lookups and (2) the point processing sequence for the randomly selected pixels in a batch.

\subsection{Developed Locality-sensitive 3D Hash Mapping Function}
\label{sec:alg_hashing}
The embedding interpolation in iNGP always fetches the embeddings of the eight surrounding vertices in the 3D cube (see Fig. \ref{fig:iNGP_training_process}). Leveraging this to enhance the locality of hash table lookups, we propose to adopt Monton code~\cite{morton}, which maps neighboring vertices in a 3D scene to neighboring hash table entries, as a locality-sensitive 3D location hash mapping function. This hash mapping function can be formulated as:
\begin{align}
h(\mathbf{x})=\left(f(x_0)+(f(x_1) \ll 1)+(f(x_2) \ll 2) \right) \bmod T 
\label{eq:zorder_hash}
\end{align}

\noindent where $T$ is the number of entries per hash table level and $f(x)$ is a separate-one-by-two function such that two zero bits are inserted between every pair of the adjacent bits (e.g., $f({\underline{1011}}_2)={\underline1}00{\underline0}00{\underline1}00{\underline1}_2$).

In this way, data locality during hash table lookups for one point's 3D cube is greatly enhanced. As shown in Fig.~\ref{fig:alg_hashing}, with Morton encoding, 82.0\% of the index distances between two neighboring vertices of one 3D cube is less than 16 entries in the hash table and none is larger than 5000; in contrast, for the original design in~\cite{muller2022instant}, only 55.4\% of neighboring vertices have index distances $\leq$16 and 22.7\% are $>$5000. Additionally, since the memory requests adopt a row-wise granularity with a commonly-used row size of 1KB~\cite{oh20143lpddr4}, our hash mapping function needs 1.58 average memory requests for one 3D cube, while the original design requires 4.02 on average.


\subsection{Proposed Ray-first Point Streaming Order}
\label{sec:alg_point_stream_order}
{\name}'s algorithm further incorporates a ray-first point streaming order, 
where points along one ray are streamed into the accelerator for processing before moving on to the next ray.
This streaming order offers two benefits.
\textbf{First}, this streaming order enhances the local register hit rates and reduces unnecessary memory requests, since neighboring points along a ray with the same surrounding cube will lookup the same embeddings (as shown in Fig.~\ref{fig:alg_stream}(a)). 
\textbf{Second}, based on the fact that neighboring points along a ray tend to have neighboring surrounding cubes, we can combine the locality-sensitive hash mapping function with the ray-first point streaming order to further enhance the 
locality of hash table lookups: 
Our evaluation shows that 
this combination leads to 3.27$\times$$\sim$35.9$\times$ effective memory bandwidth improvement (as shown in Fig.~\ref{fig:alg_stream}(b)).
\vspace{0.3em}
\section{{\name} Accelerator }
\label{sec:arch}

Our {\name} accelerator can consider one or several DRAM dies, where each bank is equipped with its own Instant-NeRF microarchitecture, as shown in Fig.~\ref{fig:lpddr}.
In this section, we first introduce {\name}'s microarchitecture per bank that integrates a mixed-precision computation logic to cope with different data types in iNGP. Then, we present our optimized hash table mapping scheme for {\name}'s algorithm. After that, we describe our heterogeneous inter-bank parallelism design, which orchestrates the heterogeneous steps with the inter-bank parallelism opportunities to minimize the costly inter-bank data movements.


\subsection{{\name}'s Microarchitecture}
\label{sec:arch_overview}
\begin{figure}[h]
  \centering
  \includegraphics[width=1.0\linewidth]{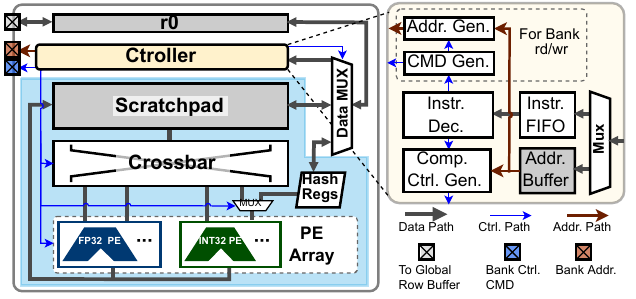}
\caption{{\name}'s microarchitecture per bank.}
\vspace{-1em}
\label{fig:hw_overall}
\end{figure}

As illustrated in Fig.~\ref{fig:hw_overall}, {\name}'s microarchitecture comprises a compute engine (in blue) and a controller (in brown). 
\ul{Compute Engine:} This engine is to compute iNGP's bottleneck steps and consists of a processing element (PE) array, a scratchpad memory, a crossbar, and hash registers for storing  pre-defined parameters of the hashing function. Specifically, the PE array consists of separate (1) INT32 PE group and (2) FP32 PE group for corresponding training arithmetics: INT32 PEs for index calculations via hash mapping function and FP32 PEs for other computations. 
The scratchpad memory feeds input data to PEs from the crossbar and stores the output data of the PEs. In addition, the INT32 PEs allow direct parameter access from the hash registers.

\ul{Controller:} The controller has two main functionalities: (1) controlling the processing of the compute engine and (2) generating read/write commands/addresses for the memory banks. It includes an instruction FIFO, an instruction decoder, an address buffer, a compute engine control signal generator, a bank command generator, and a bank address generator. Here the instruction decoder reads instructions from the FIFO and controls the other blocks to generate proper signals to implement the required functionalities. 

To read memory data into the compute engine, write data to the memory banks, or load instructions into the controller, {\name}'s microarchitecture adopts a commonly used design where a data transfer MUX is connected to each bank's global row buffer via a row-buffer sized register (i.e., r0 in Fig.~\ref{fig:hw_overall})~\cite{devic2022pim}.

\subsection{Proposed Hash Table Mapping Scheme}
\label{sec:arch_datamapping} 
\begin{wrapfigure}{r}{0.5\linewidth}
  \centering
  \includegraphics[width=0.95\linewidth]{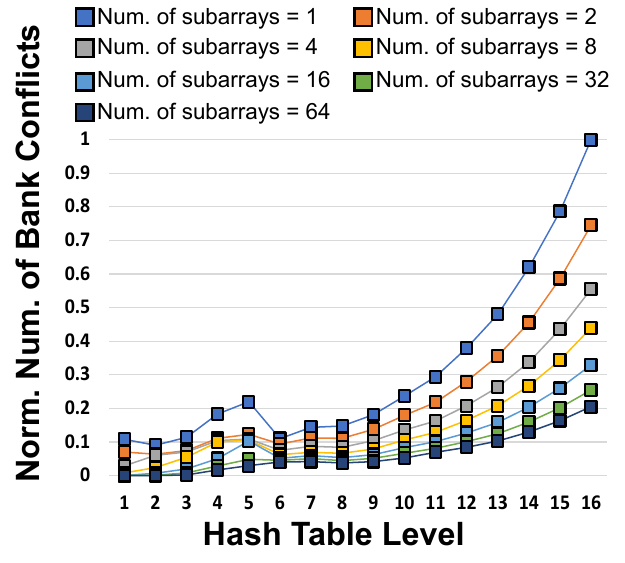}
  \caption{The normalized number of bank conflicts.}
  \label{fig:bank_conflic}
\end{wrapfigure}

\begin{figure*}[h]
  \centering
  \includegraphics[width=0.98\linewidth]{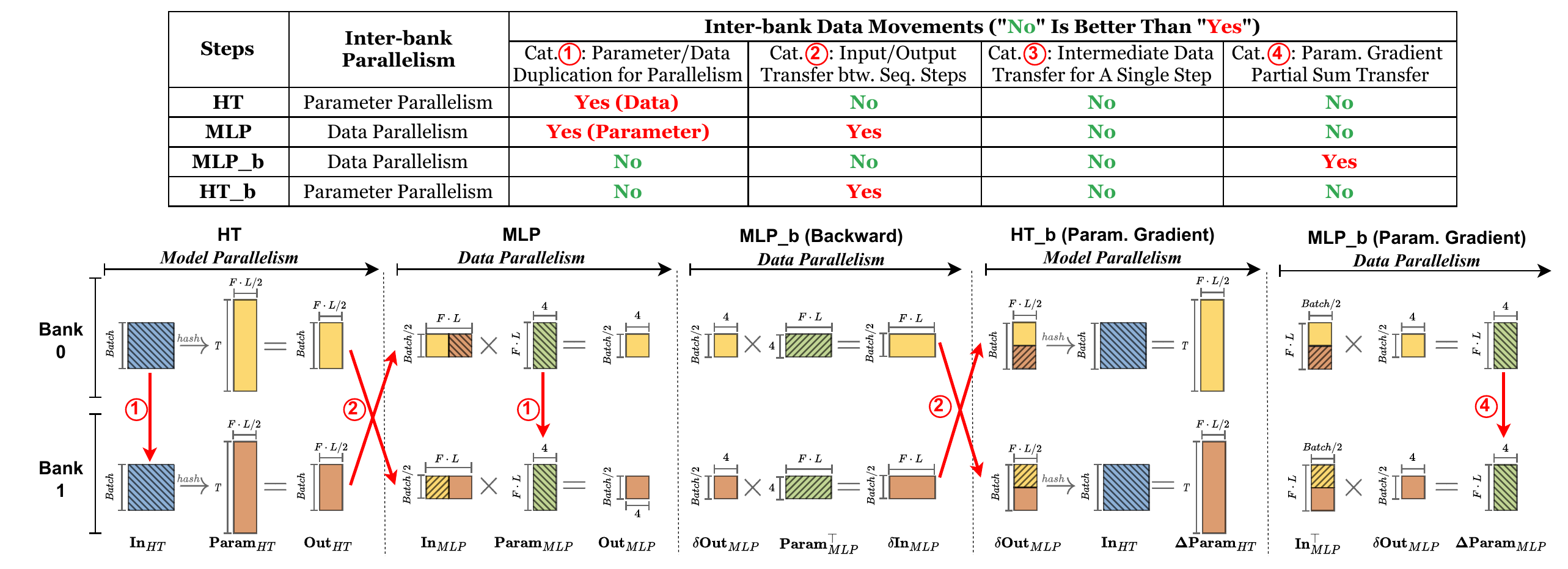}
\vspace{-0.3em}
\caption{An example that illustrates the proposed heterogeneous inter-bank parallelism design on 2 physical memory banks, i.e., \textit{parameter parallelism} and \textit{data parallelism} for HT/HT\_b and MLP/MLP\_b, respectively. Here, Cat. is short for Category, and ``Yes" and ``No" denote whether the corresponding step incurs inter-bank data movements or not.}
\vspace{-0.5em}
\label{fig:inter-bank}
\end{figure*}

To ensure satisfactory throughput, multiple points (e.g., 32 points in our evaluation) are processed in parallel in $\text{HT}$/$\text{HT}_{\_\mathbf{b}}$. Even with {\name}'s algorithmic techniques, bank conflicts due to the random hash table lookups can still cause processing stalls. 
To further mitigate bank conflicts, we develop an optimized hash table mapping scheme that leverages subarray parallelism. Our mapping scheme is divided into \textit{intra-level hash table mapping} and \textit{inter-level hash table mapping}. \ul{Intra-level Hash Table Mapping:} Leveraging the statistics that $>$50\% of the bank conflicts for one hash table level are incurred by memory requests with sequential addresses, we rearrange the sequential addresses to multiple subarrays. 
This allows these memory addresses to be requested in parallel, avoiding bank conflicts.
\ul{Inter-level Hash Table Mapping:} Fig.~\ref{fig:bank_conflic} shows the normalized number of bank conflicts for the 16 hash table levels after adopting the proposed intra-level hash table mapping scheme. We can observe that the processing time of different levels is unbalanced due to the unbalanced bank conflicts. 
To alleviate the accelerator resource under-utilization caused by these unbalanced processing times, 
we further adopt inter-level hash table mapping, where $Levels$~0$\sim$4, $Levels$~5$\sim$8, and $Levels$~9$\sim$10 are clustered into three groups. We further distribute these three groups and the other levels to different memory banks for balancing the overall processing time.

\subsection{Proposed Heterogeneous Inter-Bank Parallelism Design} 
\label{sec:arch_parall}

\begin{table}[b]
\vspace{-1.em}
\caption{Parameter/data sizes for iNGP's bottleneck steps.} 
\begin{threeparttable}
\centering
\resizebox{0.85\linewidth}{!}{
\begin{tabular}{|c|c|c|c|c|}
    \hline
    \multirow{2}{*}{\textbf{Steps}} & \multirow{2}{*}{\textbf{Param.}$^{\triangleleft}$} & \textbf{Input} & \textbf{Output} & \textbf{Intermediate} \\ 
    &                & \textbf{Data$^{\diamond}$} & \textbf{Data$^{\diamond}$} & \textbf{Data$^{{\diamond},{\dagger}}$} \\ \hline
    $\text{HT}$      & 25MB    & 3MB   & 16MB  & 0 \\ \hline
    $\text{MLP}$$^*$        & 0.014MB & 16MB  & 1.5MB & 32MB \\ \hline
    $\text{MLP}_{\_\mathbf{b}}$$^*$     & 0.014MB & 1.5MB & 16MB  & 32MB \\ \hline
    $\text{HT}_{\_\mathbf{b}}$   & 25MB    & 16MB  & 0     & 0 \\ 
    \hline
\end{tabular}
}
\end{threeparttable}
\begin{tablenotes}
\footnotesize
\item{$^*$: $\text{MLP}$ stands for applying $\text{MLP}_{\mathbf{d}}$ and $\text{MLP}_{\mathbf{c}}$ sequentially.}
\item{$^{\triangleleft}$: The multiresolution hash table size and the two MLPs' weight size for HT/HT\_b and $\text{MLP}$/$\text{MLP}_{\_\mathbf{b}}$, respectively.}
\item{$^{\diamond}$: For a batch size of 256k sampled points.}
\item{$^{\dagger}$: The max intermediate data when doing level-by-level hash table lookups or layer-by-layer $\text{MLP}$ processing.}
\end{tablenotes}
\label{tab:data_size}
\end{table} 

There are typically two approaches for designing inter-bank parallelism: (1) \textit{data parallelism} where each memory bank duplicates the parameters and processes different input data in parallel and (2) \textit{parameter parallelism} where each bank keeps a part of the parameters and performs a fraction of computations based on the same inputs duplicated across banks. Due to the limitations imposed by the I/O interface and the internal prefetch width (see Fig.~\ref{fig:lpddr}), the memory latency for accessing data from other banks is much higher than from the local bank. Therefore, \textbf{minimizing data movement sizes across different banks} is critical for maximizing the overall efficiency. 
We classify the causes of inter-bank data movements as \ul{four categories}: Category~\ding{172}~parameter/data duplication due to the adopted parallelism approaches, Category~\ding{173}~input/output data transfer between sequential steps, Category~\ding{174}~intermediate data transfer within a single step, and Category~\ding{175}~parameter gradient partial sum transfer for gradient accumulations.
Tab.~\ref{tab:data_size} illustrates the parameter and data sizes of the bottleneck steps in iNGP training. 
Based on the causes of inter-bank data movements and different data sizes of these steps,
we propose a \textbf{heterogeneous inter-bank parallelism} design to minimize the overall inter-bank data movements: we adopt \textit{parameter parallelism} for $\text{HT}$/$\text{HT}_{\_\mathbf{b}}$ (i.e., distributing the multi-resolution hash table to multiple banks), and leverage \textit{data parallelism} for $\text{MLP}$/$\text{MLP}_{\_\mathbf{b}}$ (we denote the sequential $\text{MLP}_{\mathbf{d}}$$\rightarrow$$\text{MLP}_{\mathbf{c}}$ as MLP hereafter).

\textbf{Proposed Inter-bank Parallelism Analysis.} Fig.~\ref{fig:inter-bank} exemplifies the bottleneck steps run on an {\name} accelerator 
with the proposed inter-bank parallelism design. 
This figure demonstrates how our parallelism design minimizes the inter-bank data movements for the four categories mentioned above. 
Firstly, the sizes of parameter/data duplication (Category~\ding{172}) are minimized by duplicating the much smaller parameters/input data, such as parameters in  $\text{MLP}$ (Tab.~\ref{tab:data_size}) and input data in $\text{HT}$.
Second, we only need one set of data transferred between sequential steps (Category~\ding{173}), e.g., the output data of $\text{HT}$ which is the input data of $\text{MLP}$. Therefore, the inter-bank movement sizes incurred in Category~\ding{173} are also largely reduced. 
Third, there is no intermediate data associated with Category~\ding{174}. 
Finally, the partial sum transfer for the parameter gradient accumulations in Category~\ding{175} is now constrained to handle only those for the small MLPs, leading to reduced inter-bank gradient movement sizes. 


\vspace{0.5em}
\section{Evaluation  }
\label{sec:evaluation}
\vspace{-0.5em}

\subsection{Evaluation Setup}
\label{sec:evaluation_setup}

\ul{Datasets:} Eight datasets of Synthetic-NeRF~\cite{mildenhall2020nerf}. \ul{Algorithm Baselines:} The original NeRF~\cite{mildenhall2020nerf} and three SOTA NeRF training methods~\cite{muller2022instant,garbin2021fastnerf,chen2022tensorf}. \ul{Hardware Baselines:} Two SOTA edge GPU baselines, XNX~\cite{xnx} and TX2~\cite{tx2}, whose specifications are shown in Tab.~\ref{tab:GPU_sepc}.
\ul{Implementation:} We implement the {\name} microarchitecture in RTL; synthesize it with Design Compiler; and design the layout using Cadence Innovus based on a commercial 28nm CMOS technology. {\name} layout only uses 3 metal layers since DRAM die usually has 3 metal layers. The timing and power information of {\name} microarchitecture are derived from the post-layout simulation, which is further used to simulate the whole {\name} accelerator with DRAM. 
\ul{Configuration:} Tab.~\ref{tab:eval_config} summarizes the configuration. We implement the {\name} accelerator using one DRAM die. 
\ul{Evaluation Methodology:} We build a cycle-accurate simulator extended from Ramulator~\cite{kim2015ramulator} to derive the timing and power results. 

\subsection{Algorithm Evaluation}
\label{sec:eval_algo}

For verifying the performance of our {\name} algorithm, we compare the PSNR scores (the higher the better) of SOTA efficient NeRF training algorithms and ours in Tab.~\ref{tab:algs_benchmark}. 
On average, our {\name} algorithm achieves 0.76$\sim$2.86 higher PSNR than the baselines other than iNGP. Compared with iNGP, our proposed algorithm only degrades the average PSNR by 0.23. Nonetheless, our algorithm boosts the training efficiency by 1.15$\times$ on commercial 2080Ti GPU~\cite{2080ti}.

\subsection{Hardware Evaluation}
\label{sec:eval_hardware}

\begin{table}[t]
\caption{{\name}'s accelerator parameters.} 
\begin{threeparttable}
\centering
\resizebox{0.95\linewidth}{!}{
\begin{tabular}{|c|l|c|l|}
    \hline
    \multicolumn{4}{|c|}{\textbf{DRAM Configuration}\cite{oh20143lpddr4,xnx}} \\ \hline
    \multirow{4}{*}{Timing}& \multicolumn{3}{l|}{LPDDR4-2400} \\
    &  \multicolumn{3}{l|}{LCL-tRCD-tRPpb: 4-4-6} \\
    &  \multicolumn{3}{l|}{tRAS=9, tCCD=8, tRRD=2, tRCD=4} \\
    &  \multicolumn{3}{l|}{tFAW=9, tWR=6, tRA=2$^*$, tWA=7$^*$} \\ \hline
    \multirow{6}{*}{Organization}& \multicolumn{3}{l|}{16GB total capacit} \\
    & \multicolumn{3}{l|}{128-bit I/O interface, 16-bit I/O interface/channel} \\
    &  \multicolumn{3}{l|}{8 channels, 1 rank/channel} \\ 
    &  \multicolumn{3}{l|}{1 chip/rank, 16 physical banks/chip} \\ 
    &  \multicolumn{3}{l|}{1-2-4-8-16-32-64 subarrays/bank$^*$} \\ 
    &  \multicolumn{3}{l|}{1KB local$^*$/global row buffer} \\ \hline
    \multicolumn{4}{|c|}{\textbf{{\name} Microarchitecture Configuration per Bank}} \\ \hline
    Tech. & 28nm & Frequency & 200 MHz \\ \hline
    Scratchpad & \multirow{2}{*}{2KB} & Computation & 256$\times$INT32 PEs \\
    Memory &  & Resource & 256$\times$FP32 PEs \\
    \hline
\end{tabular}
}
\end{threeparttable}
\begin{tablenotes}
\footnotesize
\item{$^*$: Parameters for subarray parallelism.}
\end{tablenotes}
\label{tab:eval_config}
\end{table} 
\begin{table}[t]
\caption{Benchmark our proposed {\name} algorithm and SOTA efficient NeRF algorithms in terms of the PSNR~\cite{hore2010image} (a higher value represents better rendering quality).}
\centering
  \resizebox{1\linewidth}{!}
  {
    \begin{tabular}{c||c||cccccccc}
    \toprule
    Methods  & Avg. & Chair & Drums & Ficus & Hotdog & Lego & Materials & Mic & Ship \\
    \midrule
    NeRF~\cite{mildenhall2020nerf} &  31.01 & 33.00 & 25.01 & 30.13 & 36.18 & 32.54 & 29.62 & 32.91 & 28.65 \\
    FastNeRF~\cite{garbin2021fastnerf} & 29.90 & 32.32 & 23.74 & 27.79 & 34.72 & 32.27 & 28.88 & 31.76 & 27.68 \\
    TensoRF~\cite{chen2022tensorf} & 32.00 & 34.68 & 25.37 & 32.30 & 36.30 & 35.42 & 29.30 & 33.21 & 29.46 \\
    iNGP~\cite{muller2022instant} & 32.99 & 34.75 & 25.81 & 33.28 & 37.31 & 36.27 & 29.51 & 36.14 & 30.89 \\
    \midrule
    \textbf{Ours} & 32.76 & 34.47 & 25.69 & 33.12 & 37.06 & 35.94 & 29.33 & 35.86 & 30.61 \\
    \bottomrule
    \end{tabular}
    }
  \label{tab:algs_benchmark}
\end{table}


\begin{figure}[t]
  \centering
  \includegraphics[width=1\linewidth]{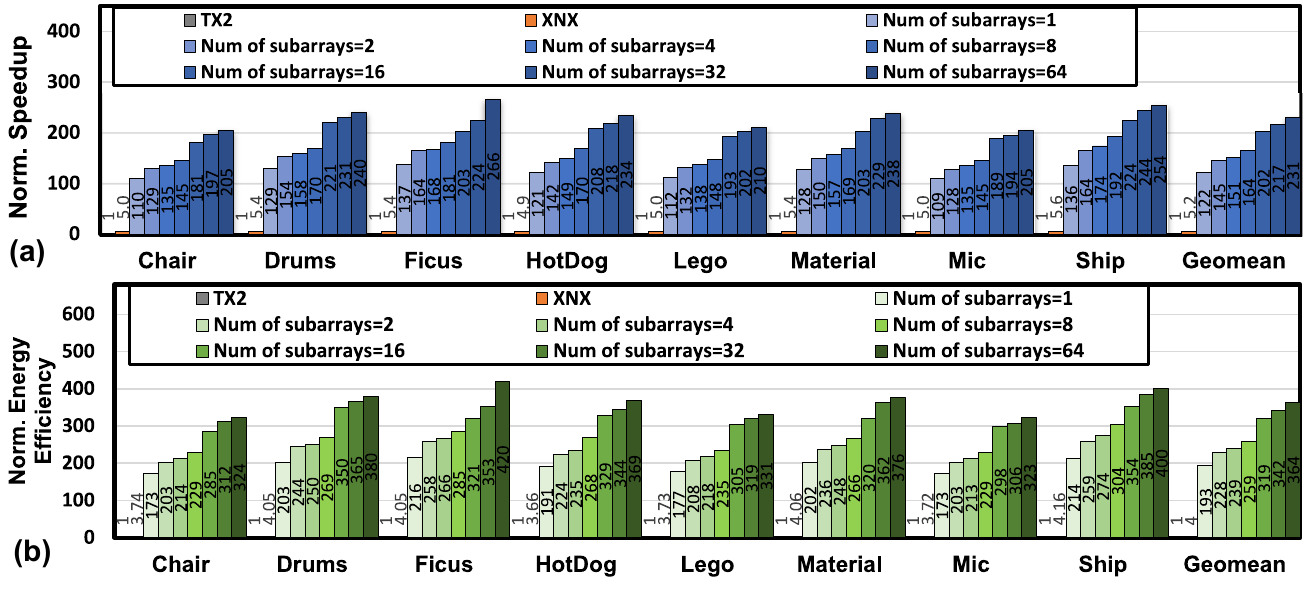}
\caption{The normalized (a) speedup and (b) energy efficiency (over TX2 GPU~\cite{tx2}) achieved by {\name} accelerator.}
\label{fig:hw_result}
\end{figure}


\ul{Area and Power:} The area of one {\name} microarchitecture is 3.6$mm^2$, which is only 1.5\% of one DRAM bank area~\cite{oh20143lpddr4}. The power of one {\name} microarchitecture is 596.3mW. 
\ul{Speedup:} Fig.~\ref{fig:hw_result}(a) presents the training time improvement achieved by the proposed {\name} accelerator in comparison with the two SOTA edge GPU baselines, i.e., TX2~\cite{tx2} and XNX~\cite{xnx}, on the eight datasets~\cite{mildenhall2020nerf}.  
Compared with the baselines, our proposed {\name} accelerator offers 109.5$\times$$\sim$266.1$\times$ and 22.0$\times$$\sim$49.3$\times$ speedup over TX2~\cite{tx2} and XNX~\cite{xnx}, respectively. 
\ul{Energy Efficiency:} Fig.~\ref{fig:hw_result}(b) presents the energy efficiency improvements. The proposed {\name} accelerator provides 172.9$\times$$\sim$420.3$\times$ and 46.4$\times$$\sim$103.7$\times$ energy efficiency improvement over TX2~\cite{tx2} and XNX~\cite{xnx}, respectively. 
\section{Related Works}
\label{sec:related}
\textbf{Near-Memory Processing.} Prior studies have utilized NMP architectures to accelerate general hash table lookups~\cite{yitbarek2016exploring} and MLP workloads~\cite{shin2018mcdram}. Our {\name} differs from prior works in that we propose an algorithm-hardware co-designed NMP framework tailored for iNGP's unique multi-resolution hash table lookups and enable dedicated inter-bank parallelisms to support iNGP's heterogeneous steps, including both hash table lookups and MLPs. 


\section{Conclusion}
We propose {\name}, the first NMP framework for enabling instant on-device NeRF training through dedicated algorithm-accelerator co-design. 
Extensive experiments on eight datasets verify that {\name} provides 22.0$\times$$\sim$266.1$\times$ speedup over SOTA edge GPUs while maintaining the rendering quality.

\section*{Acknowledgement}
This work was supported by the National Science Foundation (NSF) SCH program (Award
number: 1838873) and CoCoSys, one of the seven centers in JUMP 2.0, a Semiconductor Research Corporation (SRC) program sponsored by DARPA.




\bibliographystyle{IEEEtranS}
\bibliography{ref}

\begin{thebibliography}{10}
\providecommand{\url}[1]{#1}
\csname url@samestyle\endcsname
\providecommand{\newblock}{\relax}
\providecommand{\bibinfo}[2]{#2}
\providecommand{\BIBentrySTDinterwordspacing}{\spaceskip=0pt\relax}
\providecommand{\BIBentryALTinterwordstretchfactor}{4}
\providecommand{\BIBentryALTinterwordspacing}{\spaceskip=\fontdimen2\font plus
\BIBentryALTinterwordstretchfactor\fontdimen3\font minus
  \fontdimen4\font\relax}
\providecommand{\BIBforeignlanguage}[2]{{%
\expandafter\ifx\csname l@#1\endcsname\relax
\typeout{** WARNING: IEEEtranS.bst: No hyphenation pattern has been}%
\typeout{** loaded for the language `#1'. Using the pattern for}%
\typeout{** the default language instead.}%
\else
\language=\csname l@#1\endcsname
\fi
#2}}
\providecommand{\BIBdecl}{\relax}
\BIBdecl

\bibitem{asgari2021fafnir}
B.~Asgari \emph{et~al.}, ``{Fafnir: Accelerating sparse gathering by using
  efficient near-memory intelligent reduction},'' in \emph{27th HPCA}, 2021.

\bibitem{chen2022tensorf}
A.~Chen \emph{et~al.}, ``{TensoRF: Tensorial Radiance Fields},'' \emph{arXiv
  preprint arXiv:2203.09517}, 2022.

\bibitem{devic2022pim}
A.~Devic \emph{et~al.}, ``{To pim or not for emerging general purpose
  processing in ddr memory systems},'' in \emph{49th ISCA}, 2022.

\bibitem{morton}
C.~Ericson, Ed., \emph{Real-Time Collision Detection}, 1st~ed.\hskip 1em plus
  0.5em minus 0.4em\relax Crc Press, 2004, ch.~7, pp. 316--318.

\bibitem{garbin2021fastnerf}
S.~J. Garbin \emph{et~al.}, ``{Fastnerf: High-fidelity neural rendering at
  200fps},'' in \emph{ICCV 2021}, 2021, pp. 14\,346--14\,355.

\bibitem{hore2010image}
A.~Hore \emph{et~al.}, ``{Image quality metrics: PSNR vs. SSIM},'' in
  \emph{20th ICPR}.\hskip 1em plus 0.5em minus 0.4em\relax IEEE, 2010, pp.
  2366--2369.

\bibitem{kim2012case}
Y.~Kim \emph{et~al.}, ``{A case for exploiting subarray-level parallelism
  (SALP) in DRAM},'' in \emph{39th ISCA}, 2012.

\bibitem{kim2015ramulator}
Y.~Kim \emph{et~al.}, ``Ramulator: A fast and extensible dram simulator,''
  \emph{IEEE Computer architecture letters}, vol.~15, no.~1, pp. 45--49, 2015.

\bibitem{kwon2019tensordimm}
Y.~Kwon \emph{et~al.}, ``{Tensordimm: A practical near-memory processing
  architecture for embeddings and tensor operations in deep learning},'' in
  \emph{Proceedings of the 52nd MICRO}, 2019, pp. 740--753.

\bibitem{max1995optical}
N.~Max, ``Optical models for direct volume rendering,'' \emph{IEEE Transactions
  on Visualization and Computer Graphics}, vol.~1, no.~2, pp. 99--108, 1995.

\bibitem{workrooms}
Meta., ``{Introducing Horizon Workrooms: Remote Collaboration Reimagined},''
  \url{https://about.fb.com/news/2021/08/introducing-horizon-workrooms-remote-collaboration-reimagined/},
  2021-08-01.

\bibitem{quest_pro}
{Meta}, ``{Meta Quest Pro},'' 2022, \url{www.meta.com/quest/quest-pro/},
  2022-11-01.

\bibitem{mildenhall2020nerf}
B.~Mildenhall \emph{et~al.}, ``{Nerf: Representing scenes as neural radiance
  fields for view synthesis},'' in \emph{in ECCV 2020}.\hskip 1em plus 0.5em
  minus 0.4em\relax Springer, 2020, pp. 405--421.

\bibitem{muller2022instant}
T.~M{\"u}ller \emph{et~al.}, ``{Instant neural graphics primitives with a
  multiresolution hash encoding},'' \emph{in SIGGRAPH 2022}, vol.~41, no.~4,
  Jul. 2022.

\bibitem{tx2}
{NVIDIA}, ``{NVIDIA Jetson TX2},'' 2020,
  \url{www.nvidia.com/en-us/autonomous-machines/embedded-systems/jetson-tx2/}.

\bibitem{2080ti}
{NVIDIA}, ``{GeForce RTX 2080 TI Graphics Card},'' 2022,
  \url{www.nvidia.com/en-me/geforce/graphics-cards/rtx-2080-ti/}.

\bibitem{xnx}
{NVIDIA}, ``{Jetson Xavier NX Series 16GB},'' 2022,
  \url{www.nvidia.com/en-us/autonomous-machines/embedded-systems/jetson-xavier-nx/}.

\bibitem{oh20143lpddr4}
T.-Y. Oh \emph{et~al.}, ``{A 3.2 Gbps/pin 8 Gbit 1.0 V LPDDR4 SDRAM with
  integrated ECC engine for sub-1 V DRAM core operation},'' \emph{IEEE JSSC},
  vol.~50, no.~1, pp. 178--190, 2014.

\bibitem{shin2018mcdram}
H.~Shin \emph{et~al.}, ``Mcdram: Low latency and energy-efficient matrix
  computations in dram,'' \emph{IEEE TCAD}, pp. 2613--2622, 2018.

\bibitem{yitbarek2016exploring}
S.~F. Yitbarek \emph{et~al.}, ``{Exploring specialized near-memory processing
  for data intensive operations},'' in \emph{19th DATE}.\hskip 1em plus 0.5em
  minus 0.4em\relax IEEE, 2016, pp. 1449--1452.

\end{thebibliography}

\end{document}